\documentclass{article} 
\usepackage{iclr2015,times}
\usepackage{hyperref}
\usepackage{url}
\usepackage{graphicx}

\title{Visualizing and Comparing Convolutional Neural Networks}

\author{
Wei Yu\thanks{ This work was done when Wei Yu and Yalong Bai were interns at Microsoft Research.}, Kuiyuan Yang, Yalong Bai \& Yong Rui
\\
Microsoft Research \\
Beijing, China \\
\texttt{w.yu@hit.edu.cn} \\
\texttt{kuyang@microsoft.com} \\
\texttt{ylbai@mtlab.hit.edu.cn} \\
\texttt{yongrui@microsoft.com}\\
\And
Hongxun Yao \\
Harbin Institute of Technology \\
Harbin, China \\
\texttt{h.yao@hit.edu.cn} \\
}

\iclrconference 

\begin{document}

\maketitle

\begin{abstract}
Convolutional Neural Networks (CNNs) have achieved comparable error rates to well-trained human on ILSVRC2014 image classification task. To achieve better performance, the complexity of CNNs is continually increasing with deeper and bigger architectures. Though CNNs achieved promising external classification behavior, understanding of their internal work mechanism is still limited. In this work, we attempt to understand the internal work mechanism of CNNs by probing the internal representations in two comprehensive aspects, i.e., visualizing patches in the representation spaces constructed by different layers, and visualizing visual information kept in each layer. We further compare CNNs with different depths and show the advantages brought by deeper architecture.

\end{abstract}

\section{Introduction}
With decades of dedicated research efforts, CNNs recently made another wave of significant breakthroughs in image classification tasks, and achieved comparable error rates to well-trained human on ILSVRC2014\footnote{ILSVRC stands for ImageNet Large Scale Visual Recognition Challenge, the challenge has been run annually from 2010 to present.} image classification task~\citep{russakovsky2014imagenet}. The well-trained CNNs on ILSVRC2012 even rival the representational performance of IT cortex of macaques on visual object recognition benchmark created by ~\cite{cadieu2013neural}. CNN was introduced by~\citet{lecun1989backpropagation} for hand-written digits classification, the designed CNN architecture was inspired by Hubel and Wiesel's discovery of locally-sensitive, orientation-selective neurons in the cat's visual system~\citep{hubel1962receptive}. With several \emph{big} (in terms of number of filters in each layer) and \emph{deep} (in terms of number of layers) CNNs, ~\citet{krizhevsky2012imagenet} won the image classification competition in ILSVRC2012 by a large margin over traditional methods. The classification error rate was further significantly reduced by ~\citet{GoogLeNet,simonyan2014very,he2014spatial} with \emph{deeper} CNNs.

Though external classification behavior of CNNs is encouraging, the understanding of CNNs' internal work mechanism is still limited. In this paper, we attempt to understand the internal work mechanism by probing the internal representations (a.k.a. internal activations) in two aspects:
 \begin{enumerate}
   \item We visualize representation spaces constructed by internal layers. In CNN, each layer constructs a representation space for image patches with corresponding receptive field size. The representation spaces are visualized by t-SNE~\citep{van2008visualizing}, where patches with similar representations in a layer are showed in close positions in a 2-dimensional space.
   \item We visualize internal representations for an image via deconvolution~\citep{zeiler2014visualizing}. In CNN, each layer generates a new representation for an image in an information processing way, the new representation of each layer is projected back to the pixel space for understanding what information is kept.
 \end{enumerate}

Considering the deeper CNN designed by ~\citep{simonyan2014very} has achieved significantly better performance than the CNN used by~\citet{krizhevsky2012imagenet}, we further compare the internal work mechanism of these two CNNs (named as VGGCNN and AlexCNN respectively). The results show that VGGCNN is better at removing unrelated background information.

\begin{figure}[ht]
\centering
\includegraphics[width=0.95\textwidth,page=1]{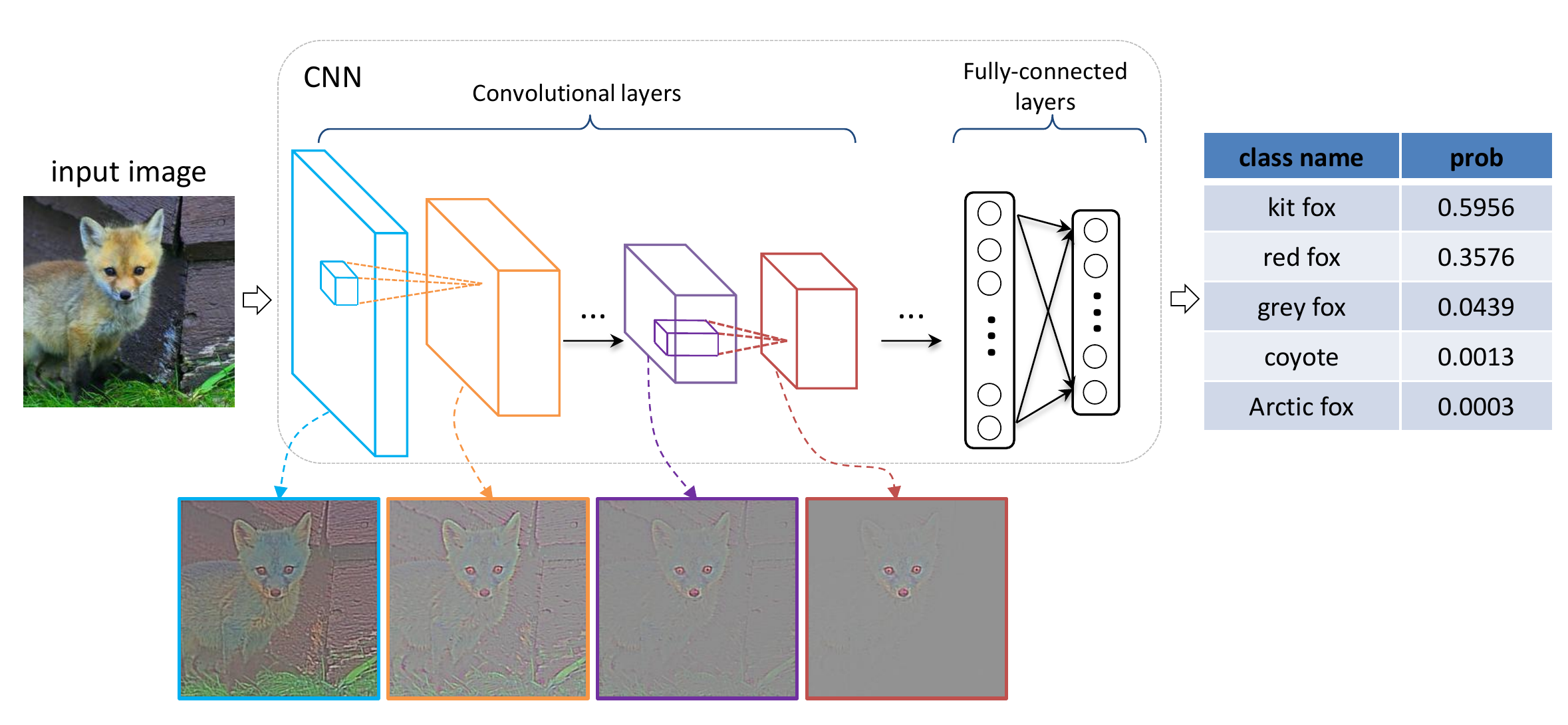}
\caption{The illustration of external and internal behavior of a CNN. The external behavior is output prediction categories for input images. The internal behavior is to be probed by visualizing the representation spaces constructed by each layer and the visual information kept in each layer.}
\label{overview}
\end{figure}

The rest of the paper is organized as follows.
We cover related work in Section~\ref{relatedWork} and then describe the architectures of VGGCNN and AlexCNN in Section~\ref{CNNConfDetail}.
The visualization of internal representations is introduced section~\ref{interMechanism}.
The comparison of VGGCNN and AlexCNN is present in Section~\ref{compCNNs}.
We discuss the conclusion in Section~\ref{conclusion}.
Section~\ref{supplementary} provides the supplementary material which visualizes the filters of VGGCNN by the method proposed by~\citet{zeiler2014visualizing}.

\section{Related work}
\label{relatedWork}
In order to open the ``black box'' of CNN, researchers have proposed several approaches to visualize the filters\footnote{In CNN, neurons are organized by layer, each neuron receives neuron activations from previous layer and weighted by weights in the connections. In fully-connected layer, each neuron is connected to all neurons in previous layers with its own weights. While in convolutional layer, neurons are further organized by feature map and only locally connected to neurons in previous layer. Moreover, all neurons in a feature map share the same filter (weights bank), so neurons in a feature map are favoring the same  kind of pattern.}to probe what kinds of patterns are these filters favoring. \citet{krizhevsky2012imagenet} directly visualized the filters learned in the first layer to judge whether the parameters of a trained CNN is apart from randomness. Since filters in high layers receive inputs from their previous layers instead of pixels, there is no direct way to visualize them in pixel space. \citet{girshick2013rich,DNNflow} used a non-parametric method, a filter is visualized by image patches with highest activations to this filter. \citet{zeiler2014visualizing} also visualize filters by patches with highest activations, together with these patches' reconstructed versions via deconvolution network. The reconstructed patch only focus on the discriminant structure in original patch, and better exhibit the filters' favored patterns.

In contrast to the above non-parametric methods, ~\citet{erhan2009visualizing} visualised deep neural networks by finding an image which maximises the neuron activation of interest by carrying out an optimisation using gradient ascent in the image space. The method was later used by~\citet{HighLevelFeatures} to visualize the ``cat'' neuron learned in a deep unsupervised auto-encoder. Recently, \citet{simonyan2013deep} employed this method to visualize neurons corresponding categories in last layer. ~\citet{InvertDNN} generalize this method to find images in the image space with similar activations in some layer to an input image.

Existing methods mostly focus on visualizing individual filter or neuron, and only partially reveal the internal work mechanism of CNN. In this paper, we do visualization in more comprehensive ways, where the representation spaces constructed by all filters of a layer are visualized, and all activations of a layers are used to reconstruct the image via deconvolution network.

\section{CNN configuration details}
\label{CNNConfDetail}
\subsection{Architecture}
In this section, we first introduce the architectures of two CNNs (AlexCNN and VGGCNN). We used the released VGGCNN\footnote{\url{http://www.robots.ox.ac.uk/~vgg/research/very_deep/}} which has achieved 29.5\% top-1 error rate on ILSVRC2012 validation set with single centre-view prediction~\citep{simonyan2014very}.
In particular, we retrain a model of AlexCNN without local response normalization layers, which achieved 42.6\% top-1 error rate with single center-view prediction~\citep{krizhevsky2012imagenet}.
Both CNNs receive RGB image with fixed size of $224\times 224$ subtracted by the mean image computed on training set.

\begin{figure}[ht]
\centering
\includegraphics[width=0.97\textwidth,page=1]{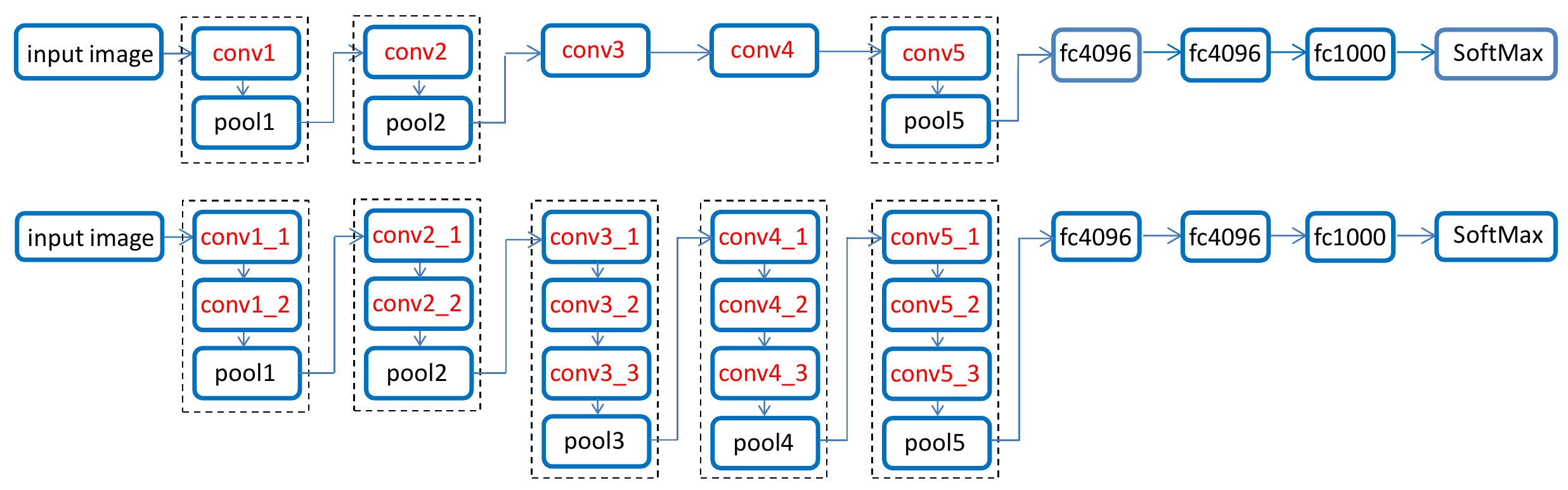}
\caption{The architectures of AlexCNN and VGGCNN. The top part is the architecture of AlexCNN, and the bottom part is the architecture of VGGCNN}
\label{architecture}
\end{figure}

The overall architectures of these two CNNs are illustrated in Figure~\ref{architecture}.
AlexCNN consists of 8 weight layers including 5 convolutional layers and 3 fully-connected layers, and three max-pooling layers are used following the first, second and fifth convolutional layers. The first convolutional layer has 96 filters of size $11\times 11$ with a stride of 4 pixels and padding with 2 pixels. The stride and padding of other convolutional layers are set as 1 pixel. The second convolutional layer has 256 filters of size $5\times 5$. The third, fourth and fifth convolutional layers have 384, 384 and 256 filters with size of $3\times 3$ respectively.

The used VGGCNN is much deeper which consists of 16 weight layers including thirteen convolutional layers with filter size of $3\times 3$, and 3 fully-connected layers. The configurations of fully-connected layers in VGGCNN are same with AlexCNN. The stride and padding of all convolutional layers are fixed to 1 pixel. All convolutional layers are divided into 5 groups and each group is followed by a max-pooling layer as showed in Figure~\ref{architecture}. Max-pooling is carried out over a $2\times 2$ window with stride 2. The number of filters of convolutional layer group starts from 64 in the first group and then increases by a factor of 2 after each max-pooling layer, until it reaches 512.

\subsection{Receptive field}
The receptive field of a neuron is its covered region in the image plane. The size and stride of receptive field of a neuron is determined by the CNN architecture. Table~\ref{VGGrf} and Table~\ref{Alexrf} list the receptive field size and stride of neurons of different layers in VGGCNN and AlexCNN, respectively.
Although both CNNs output feature maps with the same size in last pooling layer, the neurons of VGGCNN cover the receptive field with larger size.

\begin{table}
\centering
\caption{The receptive field of VGGCNN.}
\begin{tabular}{|c|c|c|c|c|c|c|c|c|c|}
\hline
layer & \textbf{c1\_1} & \textbf{c1\_2} & \textbf{p1} & \textbf{c2\_1} & \textbf{c2\_2} & \textbf{p2} & \textbf{c3\_1} & \textbf{c3\_2} & \textbf{c3\_3} \\
\hline
size & 3 & 5 & 6 & 10 & 14 & 16 & 24 & 32 & 40 \\
\hline
stride & 1 & 1 & 2 & 2 & 2 & 4 & 4 & 4 & 4 \\
\hline
layer & \textbf{p3} & \textbf{c4\_1} & \textbf{c4\_2} & \textbf{c4\_3} & \textbf{p4} & \textbf{c5\_1} & \textbf{c5\_2} & \textbf{c5\_3} & \textbf{p5} \\
\hline
size & 44 & 60 & 76 & 92 & 100 & 132 & 164 & 196 & 212\\
\hline
stride & 8 & 8 & 8 & 8 & 16 & 16 & 16 & 16 & 32\\
\hline
\end{tabular}
\label{VGGrf}
\end{table}

\begin{table}
\centering
\caption{The receptive field of AlexCNN.}
\begin{tabular}{|c|c|c|c|c|c|c|c|c|}
\hline
 layer& \textbf{c1} & \textbf{p1} & \textbf{c2} & \textbf{p2} & \textbf{c3} & \textbf{c4} & \textbf{c5} & \textbf{p5} \\
\hline
size& 11 & 15 & 47 & 55 & 87 & 119 & 151 & 167 \\
\hline
stride& 4 & 8 & 8 & 16 & 16 & 16 & 16 & 32 \\
\hline
\end{tabular}
\label{Alexrf}
\end{table}

\section{Internal work mechanism of VGGCNN}
\label{interMechanism}
In this section, we focus on visualizing the representation spaces constructed by different layers and visual information extracted in different layers.

\subsection{Representation space}
\begin{figure}[ht]
\centering
\includegraphics[width=1\textwidth,page=1]{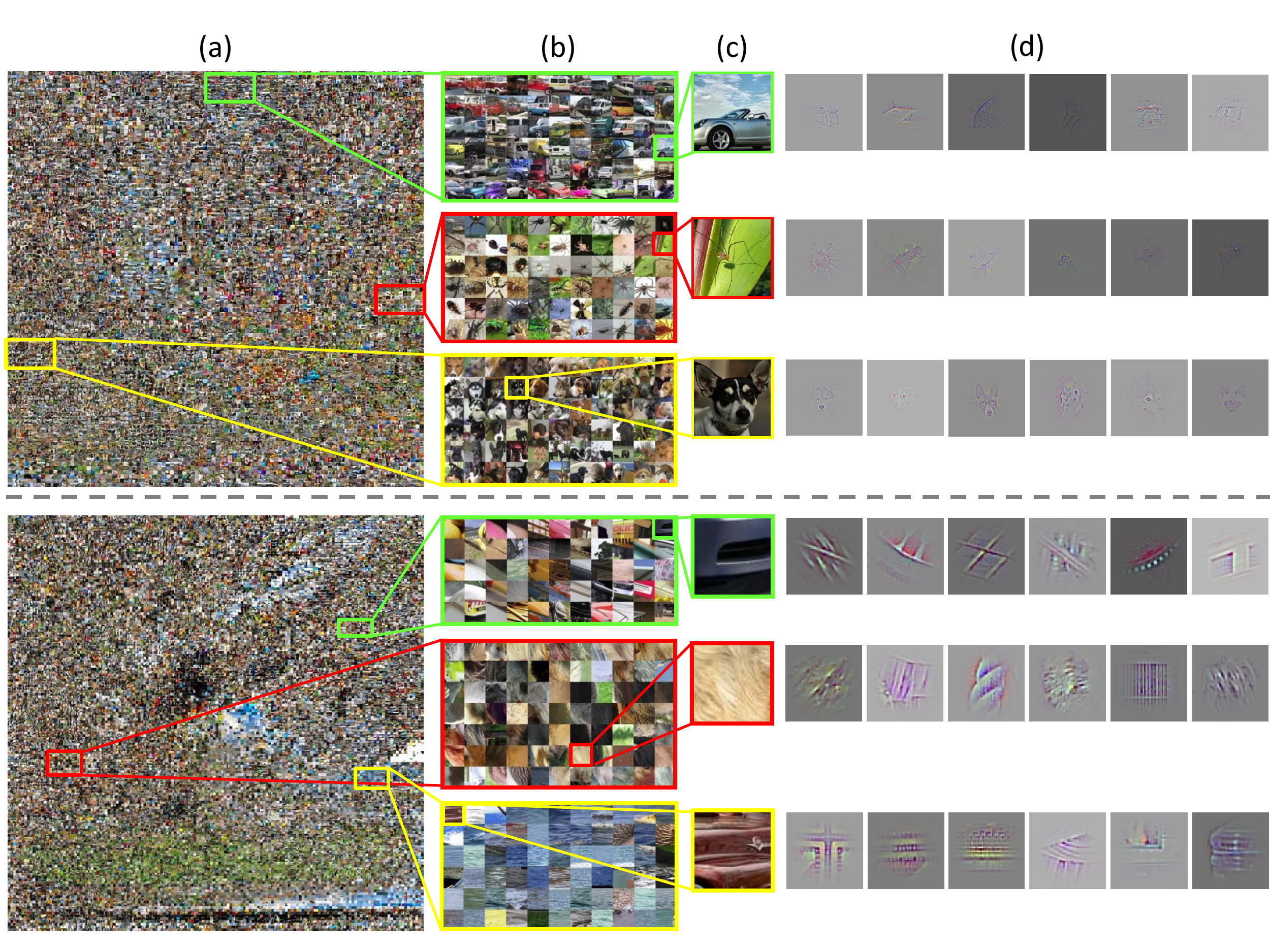}
\caption{Representation space of c5\_3 (top part) and c4\_1 (bottom part).
Column (a) is the embedding plane by dimensionality reduction, column (b) is the three subregions sampled from whole plane, column (c) is the selected patches and column (d) shows filters with high activations of corresponding patches in column (c). Filters showed by the reconstructed results of patches with highest activations in ILSVRC2012 validation set.
(Best viewed in color)
}
\label{feaSpace}
\end{figure}

As each filter generates an activation for a patch located in its receptive field, all filters in a layer actually construct a representation space for patches with size of the corresponding receptive field. Visualizing filters by their highest activated patches only partially shows each dimension of the representation space, we provide such visualization in Figure~\ref{supp} in supplementary section. Here, we utilize t-SNE~\citep{van2008visualizing} to visualize the representation space through dimension reduction, where patches close in the representation space are embedded close in the 2-dimensional space. As there are lots of empty and overlapping regions in original embedding, we fill every patch with its nearest neighbor in original embedding.
Figure~\ref{feaSpace} illustrates the representation spaces of two selected layers.

In the representation space of c5\_3, semantic-level similar patches are embedded close, e.g. the three zoom-in subregions are about \emph{car, insect, dog face}. The filters with highest activations for patches in these subregions also showed semantic-level consistence. Meanwhile, the ways of representing patches are different. In the \emph{car} example, the filters with high activations are car parts, such as car window ($1^{st}$, $5^{th}$ and $6^{th}$ filter), the part of bonnet ($2^{nd}$ filter), wheel ($3^{rd}$ and $4^{th}$ filter). In the \emph{dog} example, the filters with high activations are the dogs with different appearances or poses.

In the representation space of c4\_1, near patches are with similar texture or simple shape.
Patches in the first subregion are oblique lines or arcs. Patches in the second subregions are mainly about animal furs, while patches in the third subregion are mainly about water texture.

\subsection{Visual information extraction}
In CNN, each layer forms a new representation for an input image by gradually extracting the discriminant visual information. Here, we visualize the new representation of a layer via deconvolution network~\citep{zeiler2014visualizing} by successively reconstructing representations of previous layers until pixel space reached.
\paragraph{Reverse convolutional layer}
Representation of a convolutional layer's previous layer is reconstructed by deconvolution. Suppose convolutional layer $l$ has $K_l$ filters $\{F_k\}_{k=1}^{K_l}$ with representation organized by feature maps $\{L_k^l\}_{k=1}^{K_l}$, the reconstruction of previous layer $\{\hat{L}_c^{l-1}\}_{c=1}^{K^{l-1}}$ is formed by convolving each of the 2D feature maps $L_k^l$ with filters $\tilde{F}_{k,c}$ horizontally and vertically flipped from filters $F_{k,c}$, and summing them:$\hat{L}_c^{l-1}=\sum_{k=1}^{K_l}L_k*\tilde{F}_{k,c}$.

\paragraph{Reverse rectification layer}
Both VGGCNN and AlexCNN utilize \emph{relu} non-linearitiy following all layers, which makes all feature maps are positive.
Thus, the feature map of reconstructed layer is also ensured to be positive in reconstruction by setting negative value to zero.

\paragraph{Reverse max-pooling layer}
The locations of the maxima within each pooling window are recorded during feedforward process, and the reconstruction of previous layer is carried out by filling the recorded locations with corresponding values in current layer.

\paragraph{Reverse fully-connect layer}
Fully-connect layer can be regarded as special convolutional layer with filter size the same as input feature map. Thus, the reconstruction of fully-connect layer is the same with convolutional layer.

\begin{figure}[ht]
\centering
\includegraphics[width=1\textwidth,page=1]{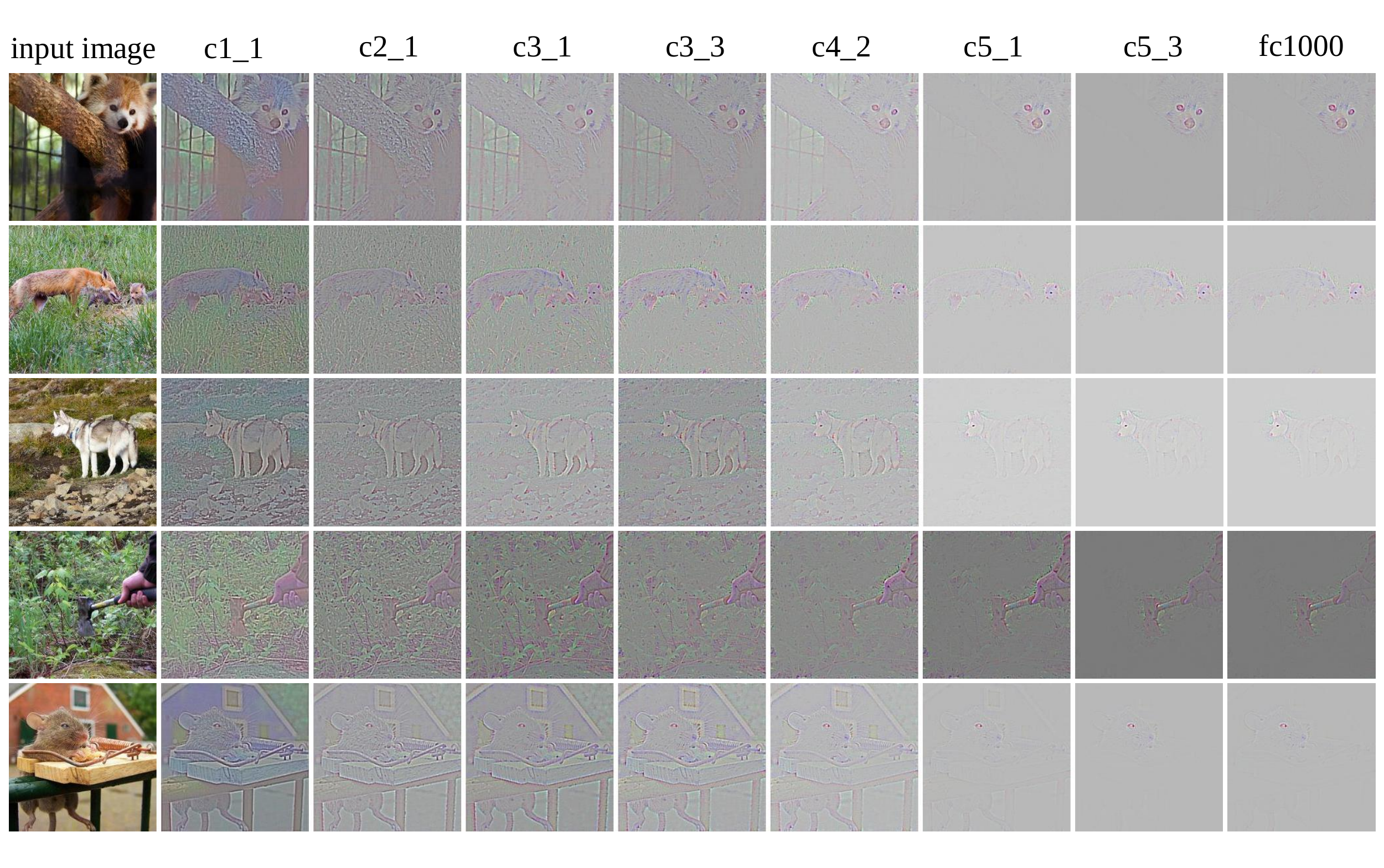}
\caption{Visualization of visual information processing through different layers.
The first column shows five input images, the following 7 columns show the reconstruction results of every two convolutional layers of VGGCNN, and the last column shows the reconstruction result of last fully-connect layer.
From top to bottom, the images are from \emph{lesser panda, kit fox, Siberian husky, hatchet} and \emph{mousetrap} respectively.
(Best viewed electronically)}
\label{visLayer}
\end{figure}

The visualization reveals the discriminant image structure that generates that representation. Figure~\ref{visLayer} shows several examples of visual results reconstructed from representations from several layers. It can be observed that, unrelated information is gradually removed from low-level layers to high-level layers (from left to right in the figure). The reconstructions of last layer only keeps the most discriminate parts. The last row shows an interesting case where mouse head is kept as discriminative part for prediction mousetrap,  this is due to mouse and mousetrap have high co-occurrence rate in images, are mouse is more discriminant in this image.

\section{Comparisons between CNNs}
\label{compCNNs}

In this section, we attempt to compare the prediction processes of VGGCNN with AlexCNN through analyzing visual information kept in different layers.

\begin{figure}[ht]
\centering
\includegraphics[width=0.95\textwidth,page=1]{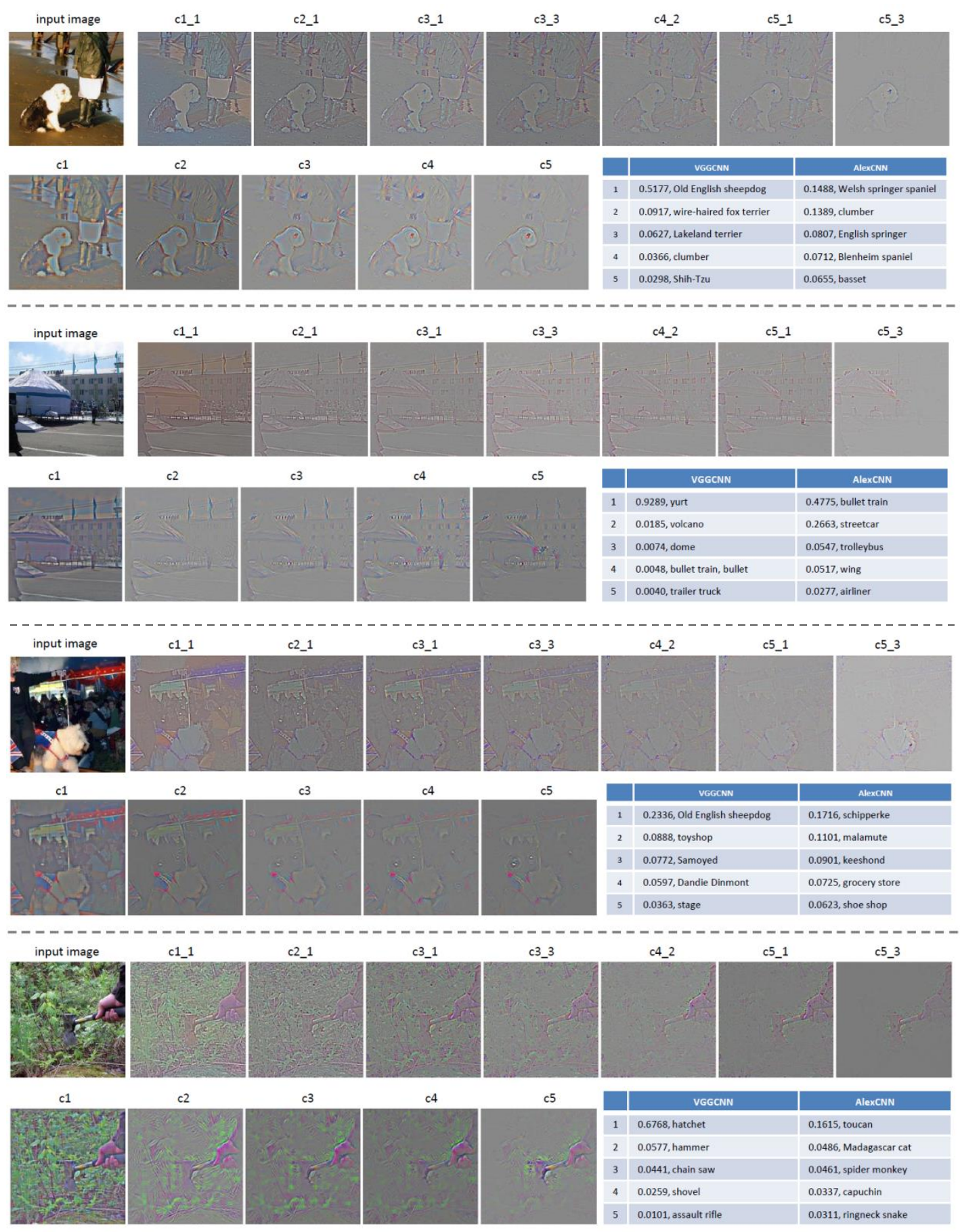}
\caption{Comparison of visual information extraction. The extraction process of VGGCNN and AlexCNN are visualized for four exemplar images. For each exemplar image, the first row show the input image followed by the reconstructed images of different layers of VGGCNN, the second row shows the reconstructed images of different layers of AlexCNN followed by the top-5 prediction results on the image.
}
\label{visComp}
\end{figure}

Previous section has shown the process of visual information extraction using visual results reconstructed from the internal representations of different layers.
It was showed that unrelated parts are gradually removed and the discriminative parts gradually stand out. In Figure~\ref{visComp}, we compare the process carried by VGGCNN and AlexCNN through illustrating several examples. In contrast to VGGCNN, AlexCNN retains more unrelated background information in last convolutional layer, which often disturbs the final prediction.

\begin{figure}[ht]
\centering
\includegraphics[width=0.95\textwidth,page=1]{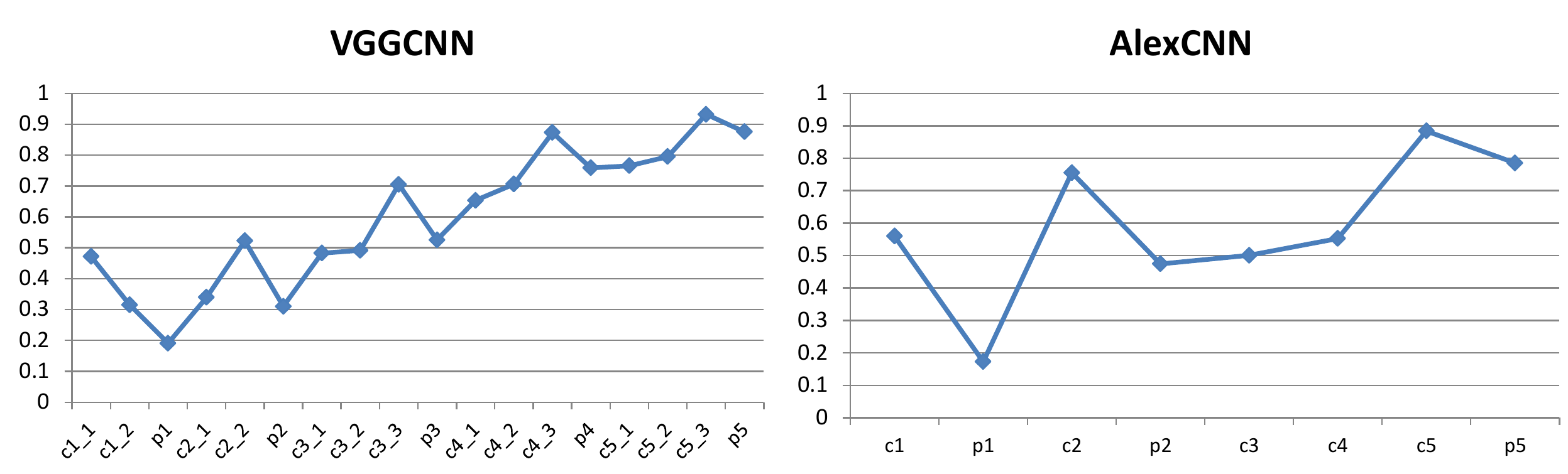}
\caption{The sparsity of each layer.
The left part is the sparsity of VGGCNN and right part is the sparsity of AlexCNN.
}
\label{sparseFea}
\end{figure}

Figure~\ref{sparseFea} shows the representation sparsity of all convolutional layers and max-pooling layers for VGGCNN and AlexCNN. The sparsity is measured by the proportion of zero activations of a layer on ILSVRC2012 validation set. In general, the sparsity increases from low-level layers to high-level layers.
To be noted that the decrease of sparsity in max-pooling layer is caused by the max operator which decreases the number of zero activations. The high-level layers of VGGCNN are with higher sparsity than AlexCNN, which demonstrates VGGCNN is with better representation ability and removing unrelated information.

\section{Conclusion}
\label{conclusion}
In this paper, we probe the internal work mechanism of CNN via visualizing the internal representations formed by different layers in two aspects. The visualizations of representation spaces constructed from different layers demonstrate the ability of CNN in sorting patterns gradually from low level to high level. The visualizations of the reconstructed images from representations of different layers show CNN's ability in gradually extracting discriminant information. Through comparison of CNNs with different depths, it shows that deeper CNN is better at extracting the discriminant information, which improves the prediction performance.

\bibliography{iclr2015}
\bibliographystyle{iclr2015}

\newpage
\section{Supplementary}
Visualization of filters from different layers of VGGCNN using deconvolution network by~\citet{zeiler2014visualizing}.
\begin{figure}[ht]
\centering
\includegraphics[width=1.0\textwidth,page=1]{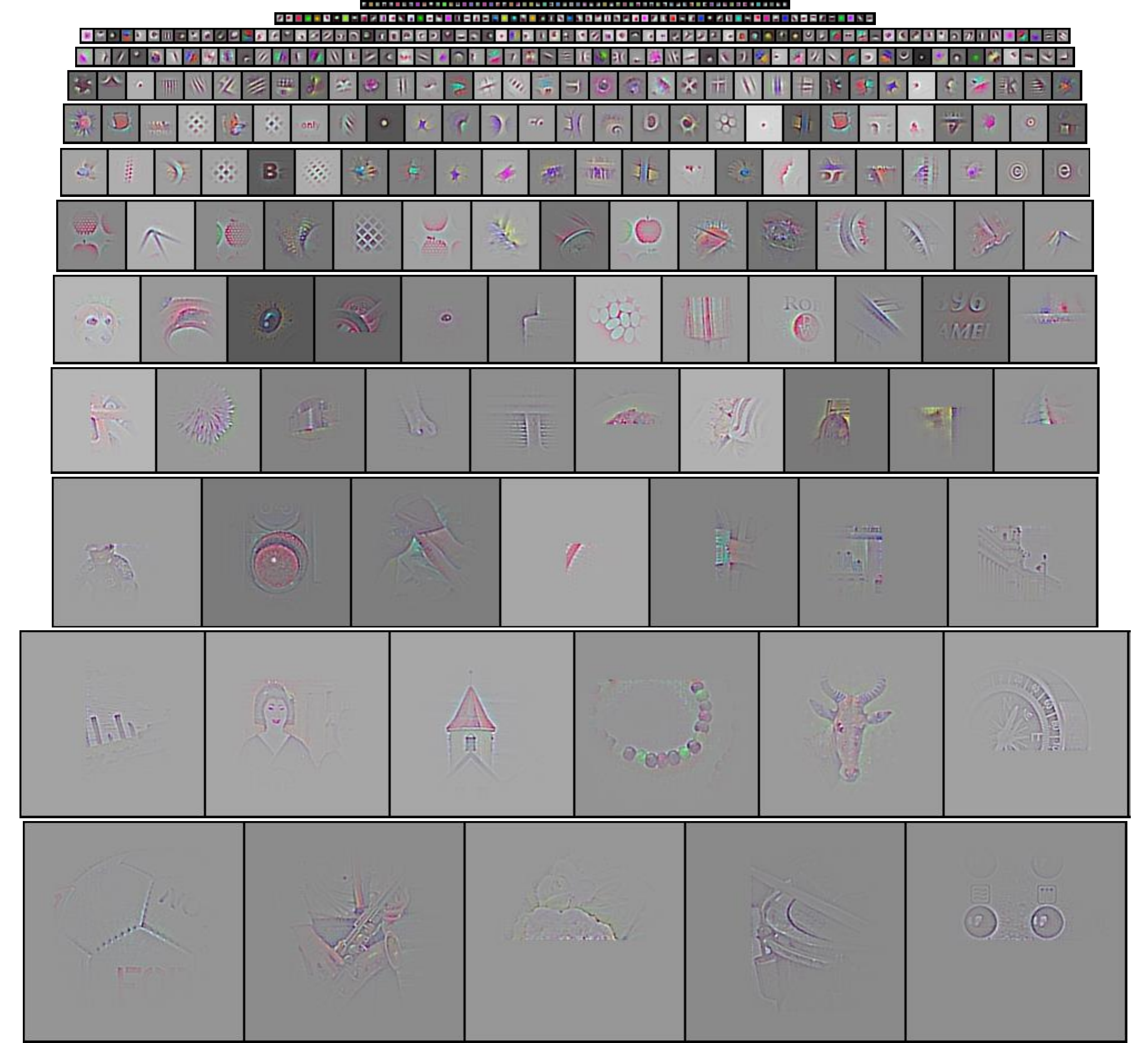}
\caption{Visualization of filters from different layers of VGGCNN. From top to down, c1\_1 to c5\_3 are showed sequentially. For layers with large receptive field size, only portion of filters are visualized. Filters in top rows are with small receptive field size and better visualized in magnified form.}
\label{supp}
\end{figure}
\end{document}